%% file: root.tex
\documentclass[letterpaper, 10 pt, conference]{ieeeconf}  % Comment this line out if you need a4paper

\IEEEoverridecommandlockouts                              % This command is only needed if 
\usepackage{caption}
\usepackage{cite}
\usepackage{dsfont}
\usepackage{amsmath,amssymb,amsfonts}
\usepackage{graphicx}
\usepackage{textcomp}
\usepackage{xcolor}
\usepackage{verbatim}
\usepackage{array}
\usepackage{tablefootnote}
\usepackage{algorithm}
\usepackage{algorithmic}
\usepackage{url}
\usepackage{wrapfig}
\usepackage{hyperref}
\usepackage{subfig}
\usepackage{booktabs}

\newcommand{\stitle}[1]{\vspace*{0.4em}\noindent{\bf #1.\/}}

\title{\LARGE \bf
$E^2DT$: Efficient and Effective Decision Transformer with Experience-Aware Sampling for Robotic Manipulation

\author{Kaiyan Zhao$^{1}$, Borong Zhang$^{2}$, Yiming Wang$^{2}$, Xingyu Liu$^{1}$, Xuetao Li$^{1}$, Yuyang Chen$^{2}$, Xiaoguang Niu$^{1}$$^{*}$}

\thanks{$^{1}$ School of Computer Science, Wuhan University, Wuhan 430072, China }
\thanks{$^{2}$ State Key Lab of Internet of Things for Smart City and Department of Computer Information Science, University of Macau, Macao 999078, China}
\thanks{$^{*}$ indicates corresponding authors: XG.Niu ({\tt\small xgniu@whu.edu.cn})} 
}

\begin{document}

\maketitle
\thispagestyle{empty}
\pagestyle{empty}

%%%%%%%%%%%%%%%%%%%%%%%%%%%%%%%%%%%%%%%%%%%%%%%%%%%%%%%%%%%%%%%%%%%%%%%%%%%%%%%%

\input{content/0_abstract}
\input{content/1_introduction}

\input{content/3_background}

\input{content/2_related_work}
\input{content/4_methodology}
\input{content/5_experiment}
\input{content/6_conclusion}

\section*{Acknowledgments}
This work was partially supported by the Key Research and Development Project of Hubei Province (2025BAB023). The calculations were performed on the supercomputing system at Wuhan University's Center for Supercomputing.

\bibliographystyle{IEEEtran}
\bibliography{references}

% \newpage
% \setcounter{theorem}{0}
% \setcounter{proposition}{0}
% \appendix
% \onecolumn
% \input{content/7_appendix}
% \onecolumn

\end{document}

%% file: content/0_abstract.tex
\begin{abstract}
In reinforcement learning (RL) for robotic manipulation, the Decision Transformer (DT) has emerged as an effective framework for addressing long-horizon tasks. However, DT’s performance depends heavily on the coverage of collected experiences. Without an active exploration mechanism, standard DT relies on uniform replay, which leads to poor sample efficiency, limited exploration, and reduced overall effectiveness.
At the same time, while excessive exploration can help avoid local optima, it often delays policy convergence and leads to degraded efficiency.
To address these limitations, we propose E$^2$DT, a DT-guided k-Determinantal Point Process sampling framework that enables the model to actively shape its own experience selection. Our framework is experience-aware, allowing E$^2$DT to be both efficient, by prioritizing sampling quality (e.g., high-return, high-uncertainty, and underrepresented trajectories), and effective, by ensuring diversity across trajectory windows to preserve policy optimality.
Specifically, DT’s internal latent embeddings measure diversity across trajectory windows, while quality is quantified through a composite metric that integrates return-to-go (RTG) quantiles, predictive uncertainty, and stage coverage (inverse frequency). These two dimensions are integrated into a novel quality–diversity joint kernel that prioritizes the most informative experiences, thereby enabling learning that is both efficient and effective.
We evaluate E$^2$DT on challenging robotic manipulation benchmarks in both simulation and real-robot settings. Results show that it consistently outperforms prior methods. These findings demonstrate that coupling policy learning with experience-aware sampling provides a principled path toward robust long-horizon robotic learning.

\end{abstract}

%% file: content/1_introduction.tex
\section{Introduction}
Autonomous decision-making and control in robotics~\cite{nguyen2019review,billard2019trends} have long been recognized as a fundamental challenge. The central goal is to enable robots to autonomously plan critical task states and execute complex tasks with high efficiency, approaching human-level capabilities. Achieving such intelligence requires robots not only to interpret the semantics of task instructions but also to infer key states from feedback signals and make informed decisions accordingly.
Among various learning paradigms, Reinforcement Learning (RL)~\cite{arulkumaran2017deep,tang2025deep} has shown remarkable promise for robotic task planning and control, allowing agents to master tasks of diverse complexity. Despite this progress, existing RL approaches still encounter major obstacles, particularly in long-term dependency modeling~\cite{ke2019modeling,cao2024deep}, exploration efficiency~\cite{ladosz2022exploration, liberty,wang2024rethinking,HuangCLLD23}, and sample utilization~\cite{zhang2021sample,yuan2024pre,bi2024sample}, which collectively hinder scalability to real-world robotic applications.

\begin{figure}[!htb]
    \centering
    \vspace{5pt}
    \includegraphics[scale=0.65]{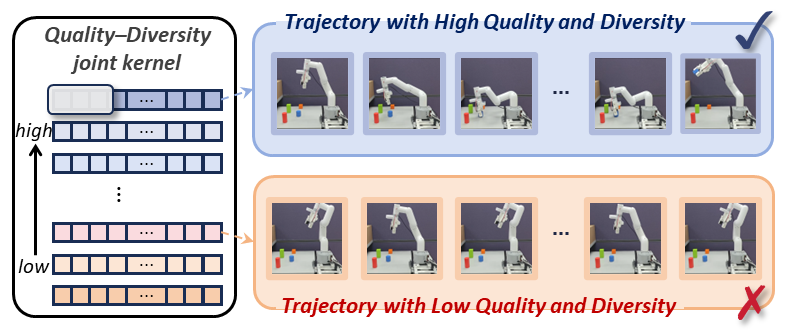}
    \caption{Our method prioritizes trajectories that jointly achieve high quality and high diversity, moving beyond simple heuristics. By scoring trajectories with a quality–diversity measure, the sampler ensures that training focuses on the most informative experiences.}
    \vspace{-10pt}
    \label{fig:e2dt_overview}
\end{figure}

To address long-horizon dependencies~\cite{lee2022dhrl,sivertsvik2024learning}, sequence modeling approaches, most notably the Decision Transformer (DT)~\cite{chen2021decision}, have emerged as a powerful paradigm. By conditioning action selection on the expected return-to-go, DT effectively captures long-horizon task structures. However, as a passive learner, its effectiveness is limited by the coverage and quality of the training data. When paired with uniform experience replay, DT suffers from redundancy, fails to prioritize high-value trajectories, and ultimately delivers poor performance.

This motivates a key insight: \emph{maximizing DT’s learning efficiency requires breaking its passive data-receiving mode and equipping it with an active, model-cooperative experience selection mechanism}. Such a mechanism must extend beyond short-sighted heuristics driven by instantaneous errors and instead evaluate the usefulness of experience from a long-term perspective. An ideal sampler should balance two orthogonal objectives: quality, meaning prioritizing experiences that directly refine the policy and carry high return potential; and diversity, meaning ensuring broad coverage of the state–action space to prevent distribution collapse and premature convergence to suboptimal solutions.

To remedy this critical deficiency, we introduce an \emph{experience-aware} framework in which the Decision Transformer actively guides its own experience sampling, jointly optimizing trajectory quality and diversity and thereby making training both \emph{efficient} and \emph{effective}.
We design a principled sampler based on a k-Determinantal Point Process (k-DPP)~\cite{kulesza2011k,kulesza2012determinantal,li2016efficientsamplingkdeterminantalpoint}, driven entirely by signals derived from the DT itself. Concretely, DT’s internal latent representations provide a rich metric space for assessing diversity, while quality is quantified by a composite of \emph{return-to-go (RTG) quantiles}, \emph{predictive uncertainty}, and \emph{stage coverage} (inverse-frequency). 
These two dimensions are unified into a novel quality--diversity joint kernel, which encourages the agent to learn preferentially from the most informative and policy-relevant experiences \emph{(driving efficiency via quality and effectiveness via diversity)} (see Fig.~\ref{fig:e2dt_overview}).
Because the k-DPP selector up-weights high-quality/rare windows, the mini-batch sampling distribution \(p_{\text{sel}}(w)\) departs from the dataset/behavior distribution \(p_D(w)\) (i.e., \(p_{\text{sel}}(w)\neq p_D(w)\)), inducing selection bias in the gradient estimates. To mitigate selection bias during training, we further adopt \emph{debiased mixed replay} with normalized importance weighting.

We term this framework \textbf{E$^2$DT} and integrate the intelligent sampling mechanism into the policy learning process to maximize data efficiency. Our contributions are as follows:
\begin{itemize}

\item We propose a learning paradigm that couples a sequence-based policy model (DT) with \emph{experience-aware} data selection, enabling the model to shape its training distribution for \emph{efficient and effective} learning.
\item We instantiate this paradigm with a DT-guided k-DPP sampler that unifies quality and diversity through a novel joint kernel, thereby balancing exploration and exploitation in the data space.
\item We conduct extensive experiments on challenging robotic manipulation benchmarks~\cite{robosuite2020} and realistic settings, demonstrating substantial improvements in sample efficiency, convergence speed, and final task success rate compared with state-of-the-art methods.
\end{itemize}

%% file: content/3_background.tex
\section{Preliminaries}

\subsection{Reinforcement Learning in Robotics}  
Reinforcement Learning (RL)~\cite{han2023survey, wang2025bile,zhao2024efficient} is a machine learning paradigm where agents learn to make optimal decisions through interactions with the environment. In robotic control~\cite{OrorbiaM23,ChenLA23,HegdeHS24}, RL is used for task planning and control, especially in complex robotic environments. The standard formulation of RL is a Markov Decision Process (MDP)\cite{gu2016deep}, represented as $ \langle \mathcal{S}, \mathcal{A}, P, R, \gamma \rangle $, where $ \mathcal{S} $ is the state space, $ \mathcal{A} $ is the action space, $ P $ defines the state transition probabilities, $ R $ is the reward function, and $ \gamma $ is the discount factor. In robotic control tasks, the state typically includes joint states, velocities, target positions, and other relevant information, while the action represents control signals, such as joint positions or velocities.
Although RL has achieved success in simpler tasks, traditional RL methods, such as DQN\cite{mnih2013playingatarideepreinforcement}, DDPG\cite{lillicrap2019continuouscontroldeepreinforcement}, and A3C\cite{mnih2016asynchronous}, face challenges in complex robotic environments, such as inefficient exploration and poor modeling of long-term dependencies. This motivates the development of new methods to improve exploration strategies and long-term decision-making.

\subsection{Decision Transformer for Long-Horizon Tasks}  
Decision Transformer (DT)\cite{chen2021decision,yuan2024transformer} is a reinforcement learning method based on the Transformer architecture, specifically designed to handle long-term reward dependencies. Unlike traditional RL methods, DT models long-term return optimization through return-to-go (RTG) and predict actions based on historical states and rewards. The goal of DT is to predict actions that maximize long-term return, conditioned on the RTG, as follows:

\begin{equation}
a_t = \text{DT}(s_t, r_t, \hat{R}_t)
\end{equation}
where $ \hat{R}_t $ represents the return-to-go starting from time step $ t $, and $ a_t $ is the action predicted by the DT model. DT excels in capturing dependencies across long time horizons, which is useful in complex robotic manipulation tasks, optimize long-term strategies and improving task success rates.

\subsection{Determinantal Point Processes (DPPs)}
A Determinantal Point Process (DPP)~\cite{kulesza2012determinantal} is a probabilistic model that favors selecting \emph{diverse} subsets. In the L-ensemble form with a positive semidefinite kernel $L \in \mathbb{R}^{N\times N}$, the probability of selecting a subset $Y \subseteq \{1,\ldots,N\}$ is
\begin{equation}
\mathbb{P}(Y) \;=\; \frac{\det(L_Y)}{\det(L + I)}\,
\end{equation}
where $L_Y$ is the principal submatrix of $L$ indexed by $Y$, and $I$ is the identity matrix of compatible size. 
Intuitively, when $L_{ij}$ encodes item similarity, $\det(L_Y)$ equals the (squared) volume spanned by the selected items’ feature vectors, larger volumes arise when items are individually salient yet mutually dissimilar, inducing negative correlations among similar items. 
In our setting, using a DPP to sample trajectory windows increases replay diversity, reduces redundancy, and improves coverage in complex environments. 
In Sec.~\ref{sec:method}, we instantiate a \textit{k}-DPP with a quality--diversity joint kernel to select informative training subsets.

%% file: content/2_related_work.tex
\section{Problem Formulation}

\begin{figure}[!htb]
\centering
\includegraphics[scale=0.60]{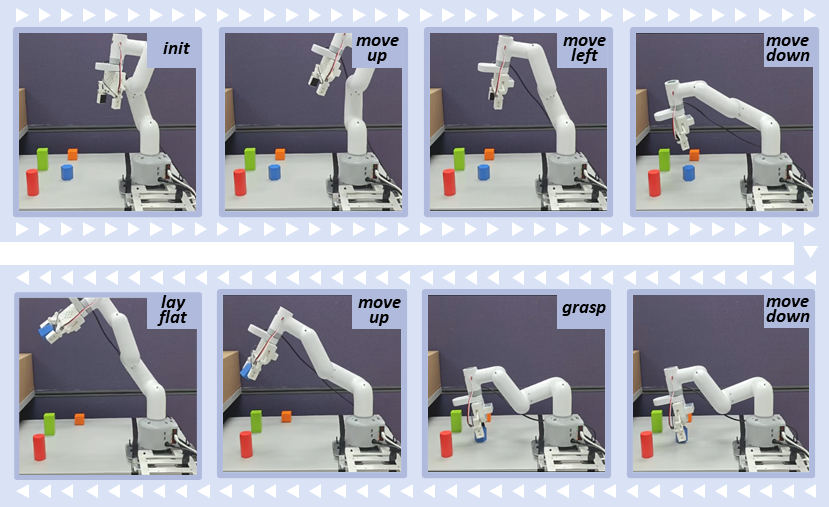}
\caption{Trajectory of a long-horizon manipulation task (Target Grasping). Success depends on sub-tasks like reaching, grasping, and rotating.}
\label{fig:ill}
\end{figure}

We begin with a typical long-horizon robotic manipulation task, such as Target Grasping, where success depends on a coherent sequence of sub-tasks like reaching, grasping, and rotating, as shown in Fig.~\ref{fig:ill}. The challenge is not only long-horizon credit assignment but also efficiently learning key operations from trajectories that contain substantial redundancy (e.g., free-space arm motions).

Our goal is to learn an efficient control policy for such tasks, modeling it as a Markov Decision Process (MDP). The policy is instantiated as a Decision Transformer (DT) \(\pi_{\theta}\) with parameters \(\theta\). In standard online learning, the agent collects new trajectories stored in a replay buffer \(\mathcal{D}\), and mini-batches are drawn uniformly for training. However, this passive scheme is limited: DT is highly sensitive to the data distribution, and uniform sampling cannot distinguish the \emph{quality} or \emph{novelty} of experience, leading to low learning efficiency.

\stitle{Core Issue: Active Experience Selection}
We argue that the challenge is not just optimizing \(\theta\), but \emph{designing an active experience selection mechanism} that cooperates with the policy \(\pi_{\theta}\). Instead of uniform sampling from \(\mathcal{D}\), we introduce a selection function

\begin{equation}
g(\cdot;\,\psi):\ \mathcal{C}\ \mapsto\ \mathcal{Y} 
\end{equation}
which selects an informative subset \(\mathcal{Y}\subset\mathcal{C}\) from a candidate pool \(\mathcal{C}\subseteq\mathcal{D}\) for policy updates (e.g., trajectory windows of length \(H\)).
This leads to the joint objective:
\begin{equation}
\label{eq:joint_opt}
\min_{\theta,\ \psi}\ \ \mathbb{E}_{\mathcal{Y}\sim g(\mathcal{C};\,\psi)}\!\left[\,\mathcal{L}(\theta;\,\mathcal{Y})\,\right]
\end{equation}
where \(\mathcal{L}(\theta;\,\mathcal{Y})\) is the training loss of the policy on the selected subset. In practice (Sec.~\ref{sec:method}), we realize \(g\) via k-DPP on a quality-diversity kernel and perform \emph{debiased mixed replay}, mixing samples from \(\mathcal{Y}\) and the global buffer \(\mathcal{D}\) with importance weighting. Equation~\eqref{eq:joint_opt} \emph{couples} policy learning (optimizing \(\theta\)) with data selection (optimizing \(g\) via \(\psi\)).

\stitle{Key Objectives}
To achieve the joint optimization in Eq.~\eqref{eq:joint_opt}, we address three key objectives:
1. Defining "High-Quality" Experience: For DT, high-quality experience balances two dimensions: (i) Quality, the utility of an experience for policy improvement (e.g., high-return trajectories or regions of high uncertainty); and (ii) Diversity, ensuring coverage of the state–action space to avoid suboptimal convergence due to distribution collapse.
2. Reliable Quality Signals: The active selector must leverage the policy’s internal signals, avoiding ad hoc heuristics, allowing DT to identify valuable experiences based on its current knowledge.
3. Balancing Optimality and Efficiency: The selection algorithm must be both theoretically principled and computationally efficient, as finding the optimal subset from a large replay buffer is intractable.

\stitle{Summary}
Our goal is to design an active experience selection framework for DT, guided by the model’s own signals, that efficiently balances quality and diversity while remaining computationally feasible, maximizing learning efficiency in long-horizon tasks.

%% file: content/4_methodology.tex
\section{Methodology}
\label{sec:method}
To address the sample inefficiency of the Decision Transformer (DT) under passive learning, we design an active experience selection framework. The core idea is to let the DT model itself guide the sampling process, thereby intelligently balancing the \emph{quality} and \emph{diversity} of training data. The framework consists of three stages: (i) signal extraction based on DT's internal states to define what constitutes high-value samples; (ii) construction of a quality--diversity joint kernel; and (iii) subset selection via k-DPP and debiased training.

\begin{figure*}[!htb]
    \centering
     \vspace{5pt}
    \includegraphics[scale=0.70]{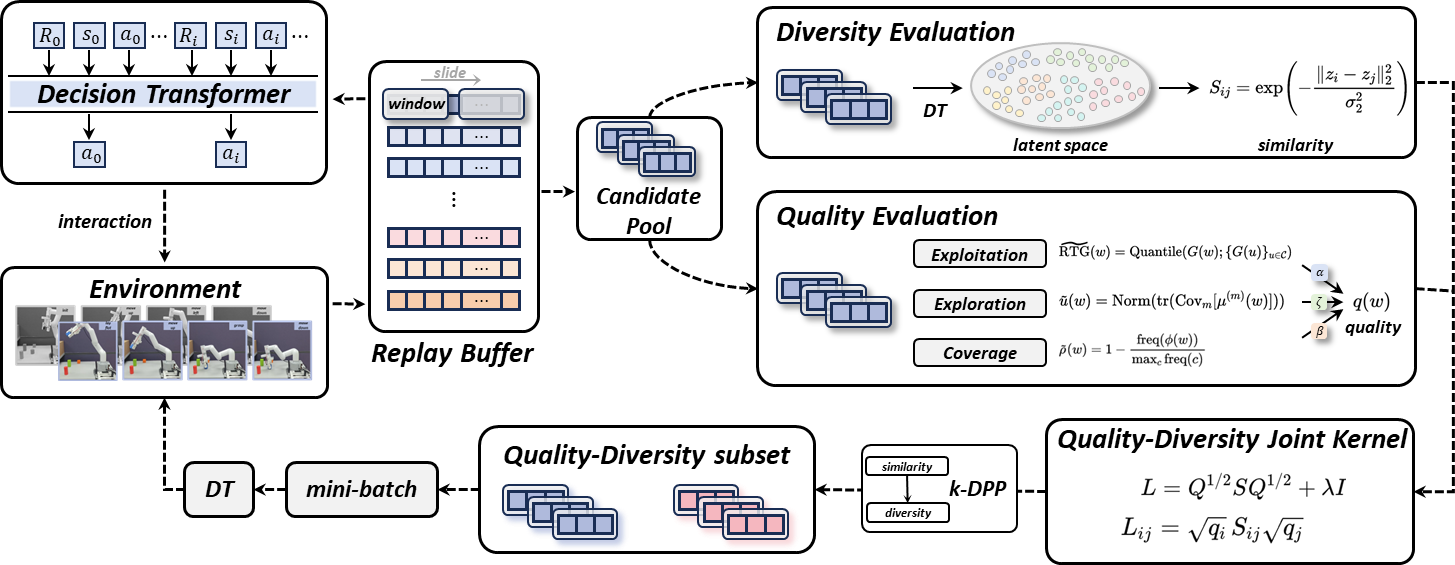}
    \caption{The E²DT framework diagram illustrates a closed-loop system that deeply integrates the decision model with active data selection. The DT model autonomously extracts internal signals from buffer samples, which are used to evaluate experiences based on two dimensions: quality (defined by return, uncertainty, and coverage) and diversity (assessed by quantifying similarity in the latent representation space between samples). These dimensions are unified into a joint kernel, which guides the k-DPP sampler to select a high-quality and diverse training subset. Finally, the Debiased Mixed Replay mechanism ensures training stability while enabling efficient learning.}
    \label{fig:framework}
    \vspace{-10pt}
\end{figure*}

\subsection{Diversity and Quality Evaluation}
The central question of this stage is: \emph{From the perspective of the current policy, what experiences are most valuable?} We posit that ideal experiences should possess both high quality (directly effective for policy improvement) and high diversity (covering a broader state--action space). To this end, we use DT's own representation capacity and predictive signals to define these two types of metrics.

\stitle{Diversity Evaluation}
To achieve effective diversity-aware sampling, we must first solve a fundamental problem: how to define and quantify the dissimilarity between trajectories. Directly computing distances in the raw physical state space (e.g., joint angles) leads to a serious pitfall: spurious diversity. For example, a robot executing the same grasp from slightly different initial poses may produce substantially different state sequences, yet remain highly redundant at the behavioral level.
\textit{Our key viewpoint is that diversity should be measured in a behavioral latent representation space rather than in the raw physical space.} This latent space is induced by the DT, where distances reflect behavioral intent rather than low-level kinematics.
To realize this, we leverage the DT model itself. The encoder $f_{\theta}$ of DT, trained to understand long-horizon context for decision making, has learned to abstract a trajectory window
\begin{equation}
 w_{t:H} \triangleq (s_{t:t+H},\, a_{t:t+H-1},\, \hat{R}_{t:t+H-1})   
\end{equation}
into a latent embedding vector $z(w)$ that captures high-level behavioral intent:
\begin{equation}
  z(w) = f_{\theta}(w) \in \mathbb{R}^d  
\end{equation}
This approach directly overcomes the deficiency of ``spurious diversity'' since $z(w)$ filters out irrelevant physical details while preserving core behavioral patterns, behaviorally similar trajectories naturally cluster in the latent representation space.
Based on this, we compute the similarity $S_{ij}$ between any two windows $i$ and $j$ in the latent space using an RBF (Gaussian) kernel:
\begin{equation}
S_{ij} = \exp\!\left(-\frac{\|z_i-z_j\|_2^2}{\sigma^2}\right)  
\end{equation}
where the bandwidth $\sigma$ can be set via the median-of-pairwise-distances heuristic. The resulting similarity matrix $S$ provides a reliable basis for the subsequent sampling algorithm to assess the true behavioral diversity within the candidate pool.

\stitle{Quality Evaluation}
To focus sampling on experiences that most improve the model, we must define what constitutes ``high quality.'' A simple criterion is high return, but this is limited because it ignores samples that, while not achieving high returns, are crucial for exploration and for remedying the model's cognitive blind spots.
\textit{Our key viewpoint is: sample quality should not be defined solely by final return, but by its marginal contribution to the current stage of learning.}
To this end, we design a multi-dimensional composite quality score $q(w)$ that balances three core pillars of learning: exploitation, exploration, and  coverage:

\begin{equation}
q(w) = \alpha \,\widetilde{\mathrm{RTG}}(w) + \beta \,\widetilde{u}(w) + \zeta \,\widetilde{\rho}(w), \; \alpha+\beta+\zeta=1   
\end{equation}

\stitle{(1) Exploitation: Identifying High-Value Samples via Return Quantiles}
This component aims to help the model quickly learn successful behavioral patterns. We first compute the discounted return of a window
\begin{equation}
 G(w) = \sum_{k=0}^{H-1} \gamma^k r_{t+k}   
\end{equation}
However, using raw returns is sensitive to outliers and scale. For more robust identification of high-value experiences, we take the \textbf{quantile} of $G(w)$ within the candidate pool $\mathcal{C}$ as a normalized score:
\begin{equation}
\widetilde{\mathrm{RTG}}(w) = \operatorname{Quantile}\!\Big(G(w);\{G(u)\}_{u\in\mathcal{C}}\Big)
\end{equation}
The advantage is that quantiles are insensitive to extreme outliers and can stably highlight samples that are ``relatively strong'' within the current candidate pool, thereby guiding the policy to converge toward validated successful directions.

\stitle{(2) Exploration: Locating Cognitive Boundaries via Uncertainty}
This component aims to guide the model to explore its ``blind spots.'' We estimate predictive uncertainty for a window by performing $M$ stochastic forward passes (e.g., MC-Dropout). Specifically, we compute the trace of the covariance of the predictive means to quantify uncertainty:
\begin{equation}
u(w) = \operatorname{tr}\!\Big(\operatorname{Cov}_m[\mu^{(m)}(w)]\Big), \; \widetilde{u}(w) = \operatorname{Norm}\big(u(w)\big)   
\end{equation}
The advantage is that large $u(w)$ precisely points to regions where the model is most uncertain and needs learning most. Prioritizing such regions is an efficient way to fill cognitive gaps and improve generalization.

\stitle{(3) Coverage: Balancing Rare Stages via Inverse Frequency}
In long-horizon tasks, data distributions are often imbalanced: common and simple behaviors (e.g., moving in free space) may overshadow rare but critical sub-tasks (e.g., final precise alignment). To address this coverage-skew problem, we introduce stage coverage. Each window is assigned a stage label $\phi(w)$, and an inverse-frequency score is computed based on its frequency in the candidate pool:
\begin{equation}
\widetilde{\rho}(w) \;=\; 1 - \frac{\operatorname{freq}(\phi(w))}{\max_c \operatorname{freq}(c)}
\end{equation}
Intuitively, $\widetilde{\rho}(w)$ converts stage frequency into a \emph{rarity weight}: the fewer windows that share the label $\phi(w)$ in the current pool, the larger the score. The mapping is bounded in $[0,1]$, monotonically decreasing in frequency, and normalized by the maximal count, so it compares stages on a common scale and is insensitive to the absolute pool size. When incorporated into the composite quality score $q(w)$, larger $\widetilde{\rho}(w)$ increases the chance that underrepresented yet pivotal sub-tasks are sampled, broadening coverage across the full task pipeline and reducing the tendency to overfit frequent trivial behaviors. In practice, stage labels can come from environment annotations or from unsupervised clustering (e.g., $K$-means) of latent embeddings $z(w)$; light count smoothing (e.g., add–$\alpha$) may be applied to avoid overemphasizing extremely rare outliers.

By fusing these three components with weights, our quality evaluation surpasses a single return-based metric and achieves a more comprehensive assessment of learning contribution that dynamically balances exploitation, exploration, and coverage.

\subsection{Constructing the Quality--Diversity Joint Kernel}

After separately quantifying the \emph{quality} of each sample ($q_i$) and the \emph{diversity} between each pair of samples (via the similarity matrix $S$), we face a core challenge: how to fuse these two distinct sources of information, one acting on individual samples (scalar scores), the other on sample pairs (similarities), into a unified mathematical structure that can be utilized by the subsequent sampling algorithm.

To this end, we construct an L-ensemble joint kernel $L$ with the following specific form:
\begin{equation}
% L = Q^{1/2} S Q^{1/2} + \lambda I, \text{where} \quad Q=\mathrm{diag}(q_1,\dots,q_{|\mathcal{C}|}) 
L \;=\; Q^{1/2} S Q^{1/2} + \lambda I,\;\text{with } Q=\mathrm{diag}(q_1,\dots,q_{|\mathcal{C}|})
\end{equation}

This formula is not a mere sum or product; it has deep geometric and probabilistic meaning.
$Q$ is a diagonal matrix whose diagonal entries are the sample-wise quality scores $q_i$. Hence $Q^{1/2}$ is diagonal with entries $\sqrt{q_i}$. $Q^{1/2} S Q^{1/2}$: when this multiplication is performed, the $(i,j)$-th entry of the resulting L-ensemble kernel $L$ becomes
\begin{equation}
L_{ij} = \sqrt{q_i} \cdot S_{ij} \cdot \sqrt{q_j}    
\end{equation}

% \stitle{Modulation of Diversity by Quality}
\stitle{Connecting Diversity and Quality: Why and How}
This result clearly reveals the core design. The ``association strength'' $L_{ij}$ between any two samples $i$ and $j$ in the new kernel is no longer their original latent-space similarity $S_{ij}$ alone, but is modulated by the geometric mean of their qualities ($\sqrt{q_i q_j}$).
If the quality of any sample is low (e.g., $q_i \to 0$), then regardless of its relation $S_{ij}$ with any other sample $j$, the association strength $L_{ij}$ will approach zero. This means low-quality samples have systematically diminished influence on the global diversity structure. If both samples have high quality, then their association strength $L_{ij}$ will be amplified.

Ultimately, this construction is crucial for the subsequent DPP (Determinantal Point Process) sampling. DPP aims to select a subset that maximizes the determinant of its kernel submatrix. Geometrically, the determinant represents the ``volume'' spanned by the vectors in the subset. In our L-ensemble kernel, this ``volume'' depends on both quality and diversity:
\textit{Effect of quality}: the sample quality $q_i$ determines the ``length'' (norm) of its corresponding vector. High-quality samples correspond to longer vectors and contribute more to expanding the volume.
\textit{Effect of diversity}: the pairwise similarity $S_{ij}$ determines the ``angle'' between vectors. The more similar two vectors are ($S_{ij} \to 1$), the smaller the angle and the smaller their contribution to the volume (as they are nearly collinear).

Therefore, when DPP maximizes the determinant on $L$, it must choose a set of samples whose vectors are long (high quality) and mutually wide-angled (high diversity). This perfectly realizes our goal of selecting a ``high-quality and complementary'' subset and provides a principled (non-heuristic) solution with solid theoretical underpinnings.

\begin{algorithm}[H]
\caption{E$^{2}$DT}
\label{alg:e2dt_concise}
\begin{algorithmic}[1]
\STATE \textbf{Initialize:} DT policy $\pi_\theta$; replay buffer $\mathcal{D}$; window length $H$; pool size $N$; subset size $k$; refresh period $K$; mix ratio $\eta$; weights $(\alpha,\beta,\zeta)$; RBF bandwidth $\sigma$; kernel regularizer $\lambda$; set $\mathcal{Y}\gets\emptyset$.
% \STATE \textbf{(Optional)} Pretrain $\pi_\theta$ on offline data;\; 
\WHILE{not converged}
  \STATE \textbf{Collect} transitions with $\pi_\theta$ and append to $\mathcal{D}$.
  \IF{$\mathcal{Y}=\emptyset$ \textbf{ or }$(\text{step} \bmod K)=0$} \label{line:refresh}
    \STATE \textbf{Candidate pool:} $\mathcal{C}\gets$ sample $N$ windows $\{w_i\}_{i=1}^{N}$ of length $H$ from $\mathcal{D}$.
    \STATE \textbf{Latent representation diversity:} encode $z_i\gets f_\theta(w_i)$;\; set $S_{ij}\gets \exp\!\big(-\|z_i-z_j\|_2^2/\sigma^2\big)$.
    \STATE \textbf{Per-window quality:} for each $w_i$:
    \STATE \quad $\widetilde{\mathrm{RTG}}_i \leftarrow$ RTG \emph{quantile} of $w_i$ within $\mathcal{C}$;
    \STATE \quad $\tilde u_i \leftarrow$ MC-dropout uncertainty
    \STATE \quad $\tilde\rho_i \leftarrow$ inverse stage frequency;\; 
    \STATE \quad $q_i \leftarrow \alpha\,\widetilde{\mathrm{RTG}}_i + \beta\,\tilde u_i + \zeta\,\tilde\rho_i$.
    \STATE \textbf{Joint kernel:} $Q\gets \mathrm{diag}(q_1,\dots,q_N)$;\; $L\gets Q^{1/2} S Q^{1/2} + \lambda I$.
    \STATE \textbf{k-DPP selection:} select $\mathcal{Y}$ via MAP inference from $L$ of size $k$.
  \ENDIF
  \STATE \textbf{Debiased mixed replay:} sample $\mathcal{B}_Y\sim\mathrm{Unif}(\mathcal{Y})$ of size $\lfloor \eta B \rfloor$ and
         $\mathcal{B}_D\sim\mathrm{Unif}(\mathcal{D})$ of size $B-|\mathcal{B}_Y|$;\; set $\mathcal{B}\gets \mathcal{B}_Y \cup \mathcal{B}_D$.
  \STATE \textbf{Importance weights:} for each $i\in\mathcal{B}$, set
         $p_i \gets \eta\,\frac{\mathds{1}\{i\in\mathcal{Y}\}}{|\mathcal{Y}|} + (1-\eta)\frac{1}{|\mathcal{D}|}$,\;
         $\omega_i \propto 1/p_i$.
  \STATE \textbf{DT update:} minimize the weighted DT loss on $\mathcal{B}$ and update $\theta$.
\ENDWHILE
\end{algorithmic}
\end{algorithm}

\subsection{k-DPP Based Sample Selection}

The goal of this stage is to select a subset from the replay buffer that is simultaneously high in quality and non-redundant in the latent representation space.

\stitle{Subset Selection}
Given the joint kernel $L=Q^{1/2} S Q^{1/2}+\lambda I$, k-DPP~\cite{kulesza2011k} assigns to each subset $Y\subseteq\mathcal{C}$ a probability proportional to $\det(L_Y)$. The determinant equals the squared volume spanned by the vectors associated with $Y$, so larger values prefer samples that are individually strong (large norms from high $q_i$) and mutually different (large angles, low redundancy from $S$). We use maximum a posteriori selection with a fixed size $k$:
\begin{equation}
\mathcal{Y}^{\star} \;=\; \arg\max_{\mathcal{Y}\subseteq\mathcal{C},\,|\mathcal{Y}|=k}\ \log\det\big(L_{\mathcal{Y}}\big).
\end{equation}
\textit{How it works in practice:} A simple greedy MAP procedure builds $\mathcal{Y}$ one item at a time by adding the candidate with the largest marginal gain. If $Y$ is the current set, the gain of adding $i\notin Y$ is
$
\Delta(i\mid Y)=\log(L_{ii}-L_{iY}\,L_{YY}^{-1}\,L_{Yi})
$
which is the conditional variance of $i$ after accounting for $Y$ (the Schur complement). This value is large when $i$ is high quality and contributes information not already explained by $Y$, and it is small for near-duplicates. With incremental Cholesky or Sherman–Morrison updates, this runs in $O(kN^2)$ time and works well because the objective is submodular.

\stitle{Debiasing via Importance Sampling}
Training only on the selected subset $\mathcal{Y}$ introduces sampling bias. We address this with Debiased Mixed Replay: draw a fraction $\eta$ from $\mathcal{Y}$ and the remaining $1-\eta$ from the global buffer $\mathcal{D}$, then apply importance weights so that the gradient matches the true data distribution. For each sampled index $i$,
\begin{equation}
p_i \;=\; \eta\, \frac{\mathds{1}\{i\in\mathcal{Y}\}}{|\mathcal{Y}|} \;+\; (1-\eta)\, \frac{1}{|\mathcal{D}|}, 
\omega_i \;=\; \frac{\tfrac{1}{|\mathcal{D}|}}{p_i}\ \ (\ \omega_i \propto p_i^{-1}\ )
\end{equation}
Finally, we update the DT by minimizing the importance-weighted loss on the mini-batch $\mathcal{B}$:
\begin{equation}
\mathcal{L}_{\mathrm{DT}}(\theta)
\;=\;
\sum_{i\in\mathcal{B}} \omega_i \sum_{j=0}^{H-1}
\Big[-\log \pi_\theta\big(a_{t+j}\mid s_{t:t+j}, \hat{R}_{t:t+j}\big)\Big]
\end{equation}
These weights make the gradient an unbiased estimator under the full data distribution, while the mix with $\mathcal{D}$ preserves a global view and stabilizes training.

%% file: content/5_experiment.tex
\section{Experiments}
\label{sec:experiments}

We evaluate the proposed E$^{2}$DT framework across RoboSuite~\cite{robosuite2020}  and ManiSkill2~\cite{gu2023maniskill2unifiedbenchmarkgeneralizable}  simulation suites and a real-world 6-DoF Elephant Robotics myCobot arm, focusing on whether DT-guided k-DPP sampling improves \emph{learning efficiency}, \emph{coverage/diversity}, and \emph{final task success}. All methods share the same offline logs for pretraining and the same online finetuning budget and reset protocol. Results are averaged over 5 random seeds with confidence intervals.

\stitle{Baselines}
We compare against the following representative methods, each using identical observation/action spaces.
\textbf{DT}~\cite{chen2021decision}: A sequence model conditioned on return-to-go (RTG) for long-horizon credit assignment; trained with uniform replay.
\textbf{HER}~\cite{andrychowicz2017hindsight}: Hindsight experience relabeling method with off-policy training.
\textbf{SynthER}~\cite{lu2023synthetic}: Synthetic Experience Replay to generate synthetic experiences for data augmentation.
\textbf{MAPLE}~\cite{nasiriany2022augmenting}: Augments RL with a library of predefined manipulation primitives for long-horizon tasks.
\textbf{Relo}~\cite{sujit2023prioritizing}: Reducible Loss (ReLo) is a sample prioritization method that ranks samples by their learnability, measured by the consistent reduction in loss over time.
\textbf{SkillTree}~\cite{wen2025skilltree}: A hierarchical framework that distills actions into a discrete skill space for long-horizon control tasks.

\subsection{RoboSuite }
\label{sec:robosuite}

\stitle{Environment}
RoboSuite~\cite{robosuite2020} is a physics-based benchmark for robotic manipulation tasks, using the MuJoCo simulator. We evaluate four long-horizon tasks: \emph{Block Stacking}, \emph{Nut Assembly}, \emph{Door Opening}, and \emph{Pick-and-Place}.Observations include robot proprioception (joint angles, velocities, gripper state) and task-specific object states. Actions are low-level joint commands or deltas of the end effector in the operational space, using the default RoboSuite controllers. Episodes terminate on success or a fixed horizon, with Success as the primary metric. 

\begin{figure}[!htb]
    \centering
    \includegraphics[scale=0.13]{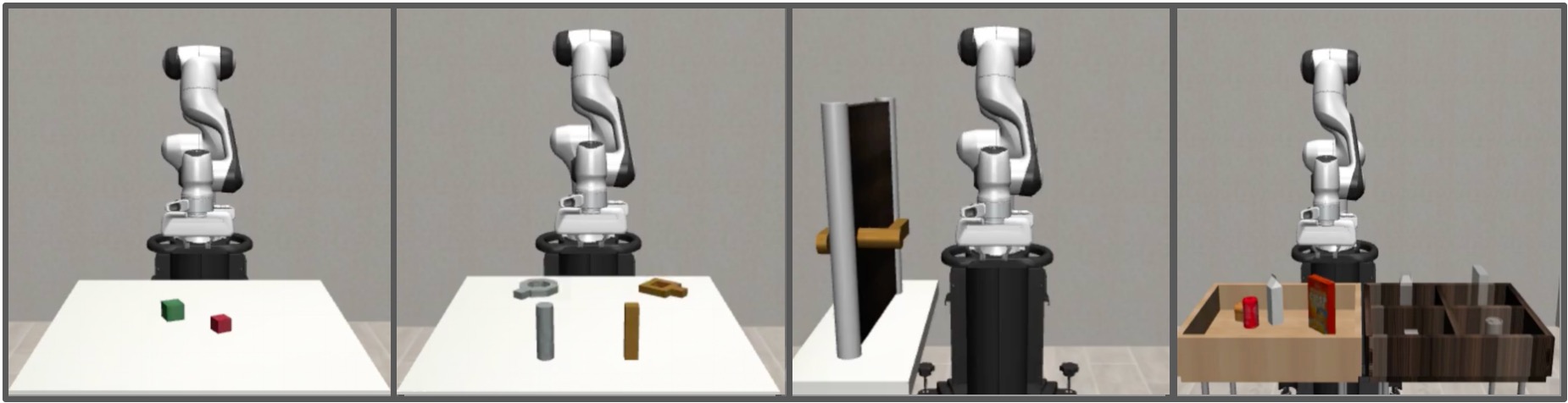}
    \caption{Manipulation Tasks:[Block Stacking, Nut Assembly, Door Opening, Pick-and-Place]}
    \label{fig:confounder_ill}
    \vspace{-5pt}
\end{figure}

We test on \emph{Block Stacking}, \emph{Nut Assembly}, \emph{Door Opening}, and \emph{Pick-and-Place}. E$^{2}$DT achieves higher final success and faster early improvement across all tasks. Gains are most pronounced in tasks with \emph{rare but critical phases} (e.g., precise alignment in \emph{Nut Assembly}, latch release in \emph{Door Opening}). Diversity in the DT latent space prevents ``spurious diversity'' in raw state space, while the quality branch (RTG quantiles + uncertainty + stage coverage) focuses training on \emph{high-value, complementary} windows, improving long-horizon credit assignment and reducing wasted exploration.

\begin{table}[ht]
\centering
\small
\setlength{\tabcolsep}{3pt}
\begin{tabular}{lcccc}
\toprule
\textbf{Method} & \textbf{BlockStack} & \textbf{NutAsm} & \textbf{DoorOpen} & \textbf{PickPlace} \\
\midrule
DT            & 60.1 $\pm$ 3.3 & 31.7 $\pm$ 3.2 & 46.8 $\pm$ 3.0 & 58.0 $\pm$ 3.4 \\
HER           & 57.0 $\pm$ 3.5 & 26.4 $\pm$ 2.9 & 41.2 $\pm$ 2.8 & 51.1 $\pm$ 3.1 \\
SynthER       & 67.2 $\pm$ 3.2 & 35.0 $\pm$ 3.0 & 49.4 $\pm$ 3.3 & 60.5 $\pm$ 2.0 \\
MAPLE         & 72.3 $\pm$ 3.0 & 42.1 $\pm$ 2.8 & 55.5 $\pm$ 3.1 & 65.6 $\pm$ 3.2 \\
Relo          & 70.5 $\pm$ 3.1 & 38.7 $\pm$ 2.9 & 53.1 $\pm$ 3.2 & 62.4 $\pm$ 2.8 \\
SkillTree     & 71.8 $\pm$ 2.9 & 40.3 $\pm$ 2.8 & 54.0 $\pm$ 2.6 & 63.5 $\pm$ 3.1 \\
\textbf{E$^{2}$DT} & \textbf{79.8 $\pm$ 2.4} & \textbf{55.6 $\pm$ 2.3} & \textbf{65.2 $\pm$ 2.4} & \textbf{73.7 $\pm$ 2.6} \\
\bottomrule
\end{tabular}
\caption{Mean success rates (\%) comparison in RoboSuite.}
\label{tab:sim_robosuite}
\vspace{-10pt}
\end{table}

\subsection{ManiSkill2}
\stitle{Environment}
ManiSkill2~\cite{gu2023maniskill2unifiedbenchmarkgeneralizable} is a high-fidelity benchmark for robotic manipulation with photorealistic rendering and physics, domain randomization for generalization.

\begin{figure}[!htb]
    \centering
    \includegraphics[scale=0.45]{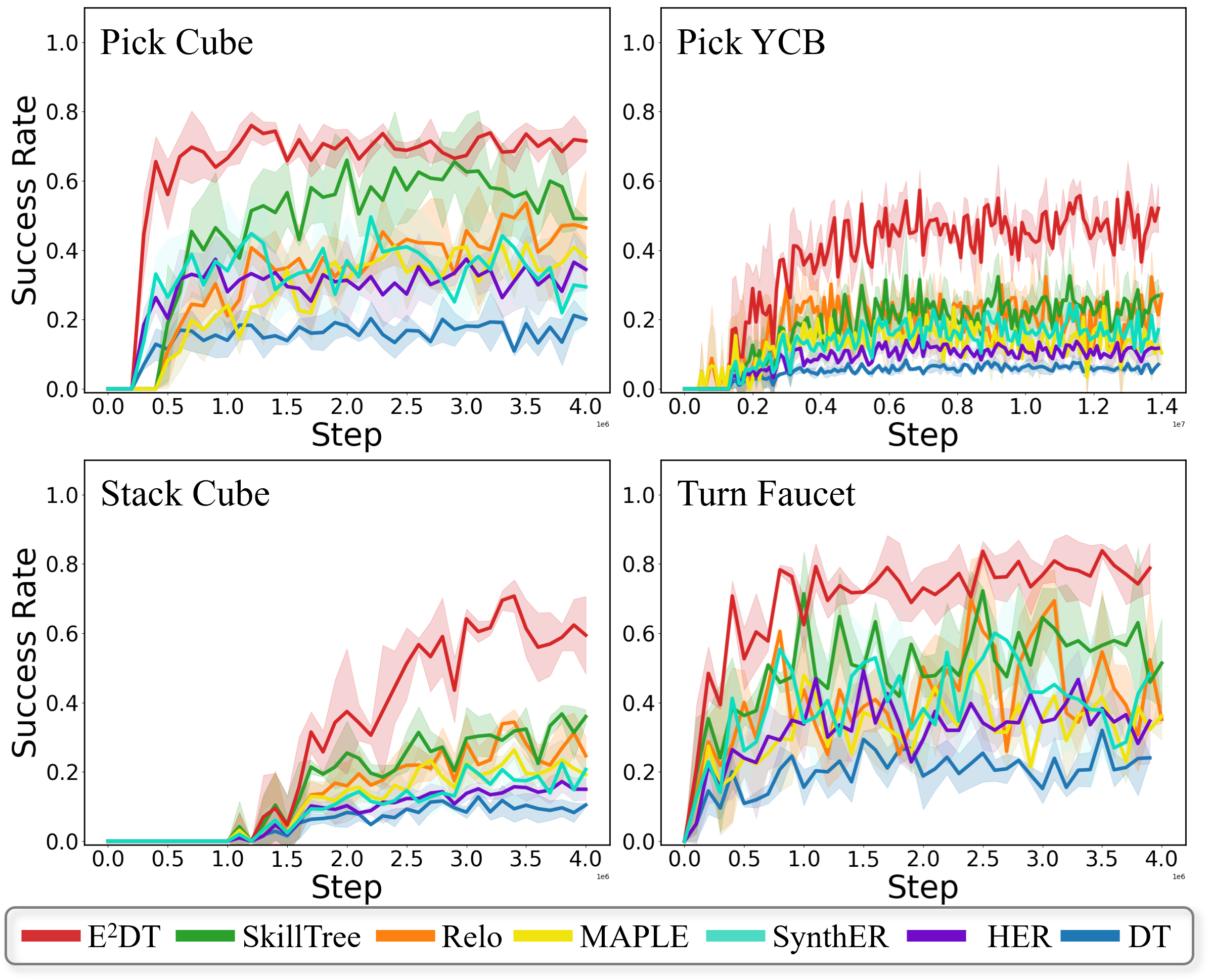}
    \caption{Mean success rates (\%) comparison in ManiSkill2.}
    \label{fig:sim_maniskill}
\end{figure}
\vspace{-5pt}

We evaluate four vision-based, long-horizon tasks: \emph{Pick Cube)}, \emph{Pick YCB}, \emph{Stack Cube}, and \emph{Turn Faucet}. Observations consist of $128 \times 128$ RGB-D images, and actions follow the benchmark's default low-level controller for the end-effector. Episodes terminate on success or a fixed horizon, with Success as the primary metric.

E$^{2}$DT achieves higher final success and faster early improvement across all four tasks (Fig.~\ref{fig:sim_maniskill}). Measuring diversity in the DT \emph{latent} embedding space avoids spurious diversity from raw pixels, while the quality composite (RTG \emph{quantiles} + predictive uncertainty + stage coverage) steers selection toward \emph{high-value, complementary} windows. Gains are most pronounced on \emph{Stack} and \emph{Turn/Open}, where rare but critical sub-stages (e.g., precise alignment, initiating rotation) dominate success; k-DPP with the joint kernel increases coverage of these sub-stages without sacrificing sample efficiency.

\subsection{Real-World Robotic Manipulation (Elephant Robotics)}
\stitle{Platform}
We deploy E$^{2}$DT on an Elephant Robotics 280 desktop manipulator (6-DoF) with a two-finger parallel gripper. The arm is base-mounted over a flat tabletop workspace. Perception uses an \emph{eye-in-hand} RGB camera on the wrist; proprioception (joint positions/velocities, gripper state) is always available. Actions are operational–space end-effector deltas (translation/rotation + gripper open/close) tracked by the vendor controller. Episodes terminate on success or a fixed horizon; standard safety interlocks (velocity limits, E-stop) are enabled.

\begin{figure}[H]
\vspace{-10pt}
  \centering
  \subfloat[High-Shelf]{%
    \includegraphics[height=2.5cm]{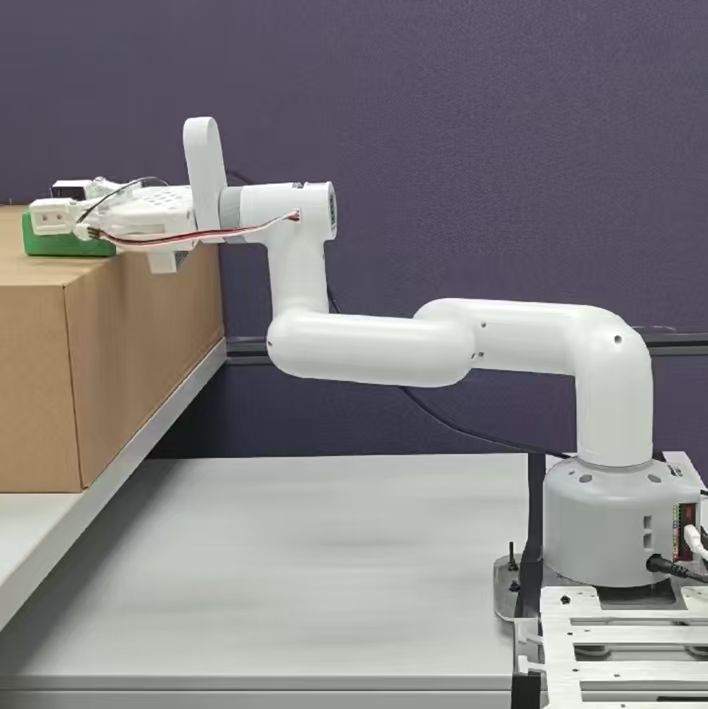}}
  \hfill
  \subfloat[Stacking]{%
    \includegraphics[height=2.5cm]{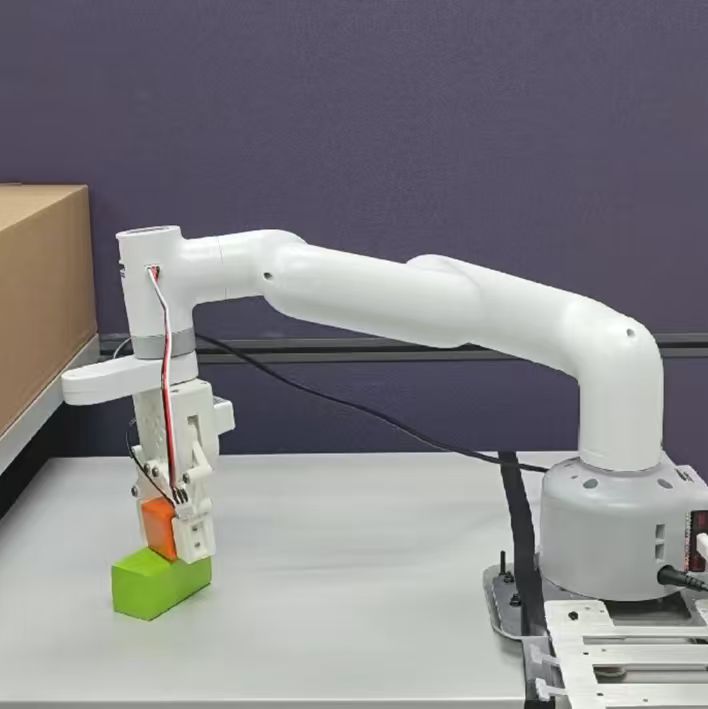}}
  \hfill
  \subfloat[Target Grasping]{%
    \includegraphics[height=2.5cm]{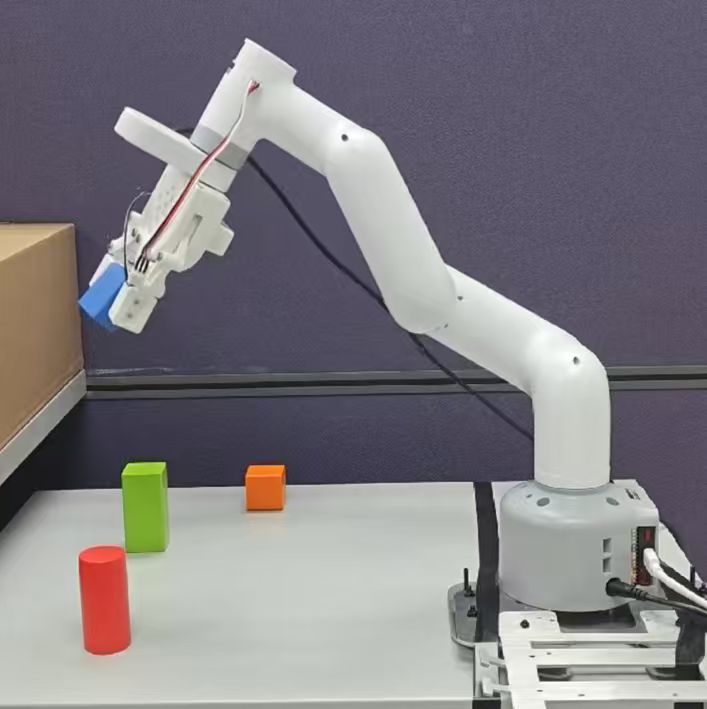}}
  \caption{Real-world tasks on Elephant Robotics 280: (a) High-Shelf Placement, (b) Stacking, (c) Target Grasping.}
  \label{fig:real_tasks}
\vspace{-5pt}
\end{figure}

\stitle{Tasks}
We evaluate three long-horizon tasks:
(i) \emph{High-Shelf Placement}: grasp an object and place it onto an elevated shelf with tight final pose tolerance;
(ii) \emph{Stacking}: sequentially grasp and stack blocks to a target configuration;
(iii) \emph{Target Grasping}: selectively grasp a specified item among distractors and hold it stably for a dwell time.

\begin{table}[H]
\centering
\small
\setlength{\tabcolsep}{6pt}
\begin{tabular}{lccc}
\toprule
\textbf{Method} & \textbf{High-Shelf} & \textbf{Stacking} & \textbf{Target Grasp} \\
\midrule
DT  & 52.4 $\pm$ 3.8 & 56.1 $\pm$ 3.2 & 54.7 $\pm$ 3.6 \\
HER  & 54.9 $\pm$ 3.6 & 53.3 $\pm$ 3.4 & 50.1 $\pm$ 3.1 \\
SynthER     & 63.2 $\pm$ 3.1 & 64.8 $\pm$ 2.9 & 61.7 $\pm$ 3.0 \\
MAPLE          & 66.5 $\pm$ 2.8 & 68.7 $\pm$ 2.7 & 66.0 $\pm$ 2.9 \\
Relo          & 61.5 $\pm$ 3.1 & 64.7 $\pm$ 3.3 & 63.0 $\pm$ 2.8 \\
SkillTree         & 69.8 $\pm$ 3.3 & 71.5 $\pm$ 2.5 & 68.9 $\pm$ 2.6 \\
\textbf{E$^{2}$DT} & \textbf{82.1 $\pm$ 2.7} & \textbf{85.6 $\pm$ 2.4} & \textbf{83.4 $\pm$ 2.5} \\
\bottomrule
\end{tabular}
\caption{Mean success rate comparison (\%, mean $\pm$ CI over 5 seeds) in real-world manipulation tasks.}
\label{tab:real_robot}
\vspace{-10pt}
\end{table}

\stitle{Results}
E$^{2}$DT achieves target success with \emph{fewer episodes} and demonstrates \emph{faster early learning} compared to baselines across all tasks. The quality branch (RTG \emph{quantiles}, predictive uncertainty, and stage coverage) focuses updates on \emph{key phases} (e.g., grasp onset, lift-to-transport transition, final alignment for shelf placement), while DT’s latent space diversity prevents “spurious diversity” from raw images. The k-DPP applied to the joint kernel $L = Q^{1/2} S Q^{1/2} + \lambda I$ enhances coverage of \emph{rare but critical} sub-stages without increasing redundancy, resulting in quicker convergence and higher final success (Tab.~\ref{tab:real_robot}).

\subsection{Ablation Study and Sensitivity}
We conduct ablations on \emph{Nut Assembly}, a long-horizon task with a rare but critical alignment phase, where data selection strongly affects learning. We evaluate four variants:
\textbf{(1) E$^{2}$DT (Full)}, k-DPP with the quality-diversity joint kernel and debiased mixed replay;
\textbf{(2) Quality-Only}, remove diversity (no k-DPP), greedily select top-$q(w)$ windows;
\textbf{(3) Diversity-Only}, remove quality (set $q_i$ constant), select purely by DT-latent space diversity via k-DPP;
\textbf{(4) Uniform Replay}, no active selection.
Removing k-DPP increases redundancy and slows learning; quality-only collapses coverage and over-focuses on local high-return regions; diversity-only preserves breadth but under-utilizes high-value samples, hurting efficiency and convergence. Only when \emph{quality and diversity are jointly enforced} via k-DPP on $L{=}Q^{1/2}SQ^{1/2}{+}\lambda I$ do we obtain the best efficiency and robustness. Performance remains stable for $k/N \in [0.08, 0.20]$ and $\eta \in [0.6, 0.8]$. Increasing $H$ improves temporal consistency but may reduce diversity in the candidate pool. The pairwise-median heuristic for $\sigma$ is robust. The choice of $K \in [5\mathrm{k}, 15\mathrm{k}]$ strikes a balance between computation and buffer-drift tracking. Similar trends are observed for the real-robot \emph{High-Shelf Placement} task.

\begin{table}[ht]
\centering
\small
\setlength{\tabcolsep}{3pt}
\begin{tabular}{lccc}
\toprule
\textbf{Variant} & \textbf{Success \%} $\uparrow$ & \textbf{Redund.\%} $\downarrow$ & \textbf{Diversity} $\uparrow$ \\
\midrule
E$^{2}$DT (Full)     & \textbf{55.6} $\pm$ 2.3 & \textbf{10.1} $\pm$ 1.2 & 0.87 $\pm$ 0.03 \\
\quad Quality-Only   & 49.6 $\pm$ 2.8          & 32.9 $\pm$ 2.1          & 0.46 $\pm$ 0.05 \\
\quad Diversity-Only & 43.8 $\pm$ 3.0          & 13.4 $\pm$ 1.5          & \textbf{0.89} $\pm$ 0.04 \\
\quad Uniform Replay & 31.2 $\pm$ 3.4          & 42.3 $\pm$ 2.7          & 0.40 $\pm$ 0.06 \\
\bottomrule
\end{tabular}
\caption{Ablation on \emph{Nut Assembly} (mean $\pm$ CI over 5 seeds). \emph{Diversity} is normalized $\log\det(S_{\mathcal{Y}})$; \emph{Redundancy} is the near-neighbor rate in the DT embedding space.}
\label{tab:ablation_nut}
\end{table}

%% file: content/6_conclusion.tex
\section{Conclusion}
We presented E$^{2}$DT, an experience-aware approach that pairs a Decision Transformer with active data selection to make training both efficient and effective. By scoring each training window for quality (RTG quantiles, uncertainty, stage coverage) and diversity (DT latent embeddings), then selecting with k-DPP and training with debiased mixed replay, E$^{2}$DT focuses on the most informative experiences. Experiments in RoboSuite, ManiSkill2, and on a real arm confirm consistent gains over strong baselines. Future work aims to automate quality–diversity weighting.